\documentclass[conference]{IEEEtran}
\usepackage{graphicx}
\usepackage{algorithm}
\usepackage{algorithmic}
\usepackage{amsmath}
\usepackage{multirow}
\usepackage{xcolor}
\usepackage[hidelinks]{hyperref} 

\ifCLASSINFOpdf
\else
\fi

\begin{document}
%
\title{Siamese-Driven Optimization for Low-Resolution Image Latent Embedding in Image Captioning}



%
\author{\IEEEauthorblockN{Jing Jie Tan\IEEEauthorrefmark{1}\IEEEauthorrefmark{2},
Anissa Mokraoui\IEEEauthorrefmark{2},
Ban-Hoe Kwan\IEEEauthorrefmark{1}, 
Danny Wee-Kiat Ng\IEEEauthorrefmark{1} and
Yan-Chai Hum\IEEEauthorrefmark{1}}
\IEEEauthorblockA{\IEEEauthorrefmark{1}Department of Mechatronics and Biomedical Engineering, Lee Kong Chian Faculty of Engineering and Science,\\ Universiti Tunku Abdul Rahman,Malaysia\\
}
\IEEEauthorblockA{\IEEEauthorrefmark{2}Laboratoire de traitement et transport de l'information, Université Sorbonne Paris Nord, France
}
Email: tanjingjie@1utar.my, anissa.mokraoui@univ-paris13.fr, \{kwanbh,ngwk,humyc\}@utar.edu.my
}


\maketitle
\begin{abstract}
Image captioning is essential in many fields including assisting visually impaired individuals, improving content management systems, and enhancing human-computer interaction. However, a recent challenge in this domain is dealing with low-resolution image (LRI). While performance can be improved by using larger models like transformers for encoding, these models are typically heavyweight, demanding significant computational resources and memory, leading to challenges in retraining. To address this, the proposed SOLI (Siamese-Driven Optimization for Low-Resolution Image Latent Embedding in Image Captioning) approach presents a solution specifically designed for lightweight, low-resolution images captioning. It employs a Siamese network architecture to optimize latent embeddings, enhancing the efficiency and accuracy of the image-to-text translation process. By focusing on a dual-pathway neural network structure, SOLI minimizes computational overhead without sacrificing performance, making it an ideal choice for training on resource-constrained scenarios.
\end{abstract}



%
\IEEEpeerreviewmaketitle
\section{Introduction}
In recent years, the field of computer vision has seen remarkable advancements, particularly in the realm of image captioning \cite{stefanini2022show}. Image captioning, which involves generating textual descriptions for visual content, has numerous applications, including accessibility for the visually impaired, content-based image retrieval, and automatic image annotation \cite{yu2023qualityagnostic}. However, the quality of captions generated for low-resolution images  remains a significant challenge due to the reduced availability of salient features and finer details \cite{Li_2024_CVPR}.

While specific research on low-resolution image (LRI) captioning may be limited, several related areas offer insights into addressing similar challenges.
In real-life scenarios, image compression is a common practice, such as on social media platforms, or data loss can occur during transmission, such as in video streaming. This process not only reduces image quality but also introduces pixel noise, posing challenges for model classification despite the images remaining human-readable \cite{yu2023qualityagnostic,imagecompress9521684,jimaging10050113,mishra2022deep}. These conditions often result in low-resolution images, posing additional challenges for image captioning systems \cite{chai2020efficient, 125072jpeg}. Hence, this is a niche area remains unexplored: the impact of LRI in image captioning. 

Several exiting strategies can be used to address this issue, including dataset augmentation, regularization techniques, specialized model architectures, and optimized loss functions tailored for low-resolution inputs \cite{ghandi2023deep}. Dataset augmentation techniques, such as image interpolation, resizing, and data synthesis, aim to improve the diversity and quality of training data. By artificially generating low-resolution variants of high-resolution images, models can better generalize across different image qualities. Studies in the field of image captioning (e.g. \cite{yu2023qualityagnostic})  demonstrated the effectiveness of data augmentation. These approaches often mimic real-life conditions, including camera blur, shadows, and long exposure effects, to enhance model robustness.

Image captioning has garnered substantial attention in recent years, with numerous approaches focused on generating descriptive text for images. Regular methods often rely on convolutional neural networks (CNNs) for image feature extraction and recurrent neural networks (RNNs) or long short-term memory networks (LSTMs) for language generation \cite{Aneja_2018_CVPR}. These models, while effective, have been increasingly supplemented and sometimes supplanted by more advanced architectures.
The introduction of transformer models, particularly visual transformers and large-scale pre-trained language models such as Generative Pretrained Transformer (GPT), has marked a significant evolution in the field. These models utilize attention mechanisms to better capture complex relationships within images and texts, offering superior performance in many cases. However, this makes retraining the model more challenging. 
Recent studies have frequently adopted ImageNet architectures such as VGG and ResNet as encoders \cite{xu2015show, show9303609}. In parallel, visual transformers have emerged as formidable alternatives in image processing \cite{castro2022deep}. Similarly, for decoders, traditional RNN and Long Short-Term Memory (LSTM) architectures have been widely used alongside newer models like GPT \cite{sharma2020image,wang2020evolutionary,vit10551257}. 

In this paper, we evaluate the performance of both conventional neural networks and transformer-based models to assess their effectiveness in generalizing to LRI. By using established scores as references and replicated model results as baselines, we aim to provide a comprehensive comparison and justify the performance of different model architectures in the context of LRI captioning. Since this research is not primarily aimed at improving previous state-of-the-arts results but rather at enabling the model to generalize to low-resolution images, we will use the scores reported in the literature as references. The results from the replicated models will serve as our baseline.

\section{Methodologies for SOLI approach}
This research consists of 4 steps in a pipeline: i) Dataset preparation and augmentation; ii) Developing the proposed model framework; iii) Training; and iv) Evaluation.

\subsection{Dataset Preparation and Augmentation}
We apply the Flickr8k dataset \cite{hodosh2013framing} for our experiments. We adapted Andrej Karpathy's training, validation, and test splits to ensure consistency \cite{Karpathy_2015_CVPR}. Table \ref{tab:flickr8k_stat} presents the statistics of image dimensions within this dataset. As shown, most of the images have similar dimensions, with the median values being close to the maximum dimensions. According to recommendations from prominent image embedding encoder such as ImageNet and the ViT model, the optimal input size is $224\times224$ pixels. We examined the impact of resizing images to this standard size (0.5 scaling factor), as well as resizing to smaller dimensions: $100\times100$ pixels (0.2 scaling factor) and $50\times50$ pixels (0.1 scaling factor). Additionally, we considered an extreme case with a 0.05 scaling factor, resulting in $25\times25$ pixels. Images resized below this threshold were excluded, as they no longer retained sufficient information.

\begin{table}[h]
    \centering
    \caption{Flickr8k Image Dimension Statistics}
    \label{tab:flickr8k_stat}
    \begin{tabular}{|c|c|c|c|c|c|}
        \hline
        \textbf{Dimension} & \textbf{Mean} & \textbf{Std Dev} & \textbf{Median} & \textbf{Min} & \textbf{Max} \\ \hline
        Height & 397.25 & 75.67 & 375.0 & 127 & 500 \\ \hline
        Width & 457.87 & 68.66 & 500.0 & 164 & 500 \\ \hline
        Channels & 3.00 & 0.00 & 3.0 & 3 & 3 \\ \hline
    \end{tabular}
\end{table}

\begin{figure*}[ht]
    \centering
    \includegraphics[width=\linewidth]{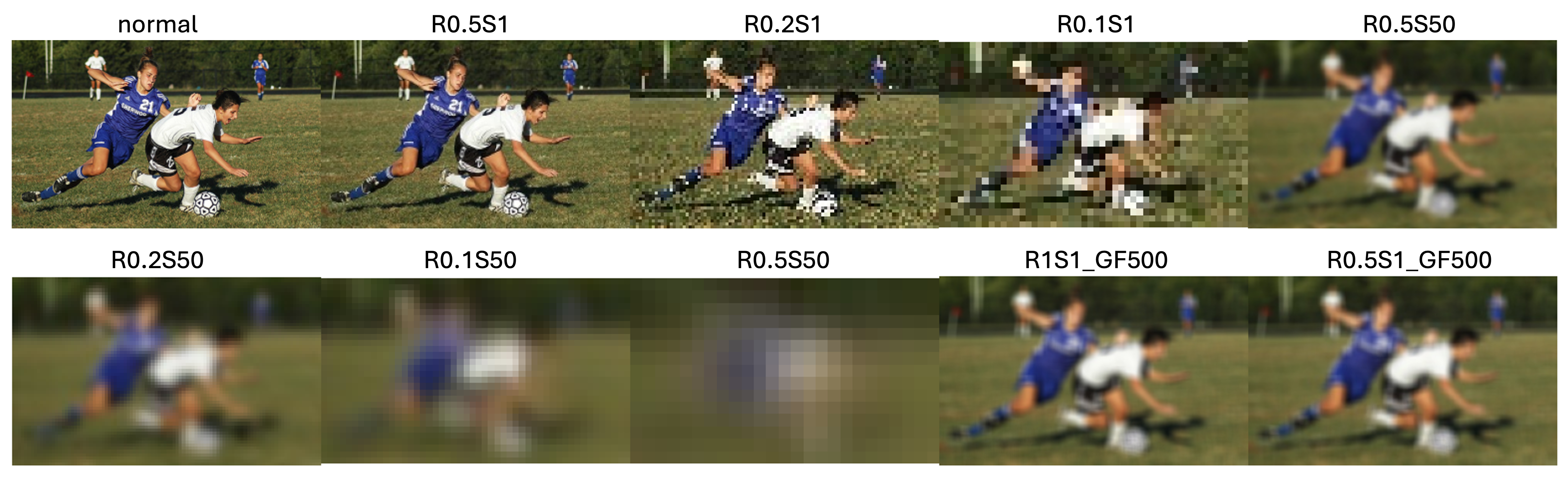}
    \caption{The sample image from the Flickr8k dataset undergoing different augmentation techniques. Here, \( R \) denotes the ratio, \( S \) denotes the step, and \( GF \) denotes the Gaussian filter's standard deviation (\(\sigma\)). For instance, \( R0.5S1\_GF500 \) indicates an augmentation with a ratio of 0.5, a step of 1, and a Gaussian filter with a standard deviation of 500. Note that for illustration purposes, the image has been resized again to a similar resolution.} 
    \label{fig:sample}
\end{figure*}

To simulate real-world image augmentations, often due to network transmission and compression by various social media platforms, we employed three variations of resizing: standard resizing, step resizing, and resizing with augmentation. The original Flickr8k dataset, referred to as the "normal" dataset. Samples of the augmented images are shown in Fig. \ref{fig:sample}.

\subsubsection{Standard Resizing}
In standard resizing, we utilized the cv2.resize algorithm from the OpenCV library \cite{opencv_library}. This algorithm is chosen for its implementation of bilinear interpolation, a widely employed method in common social media compression. Bilinear interpolation operates by estimating pixel values based on the average of the closest neighboring pixels in both dimensions during image resizing \cite{khaledyan2020low}. This method smooths transitions between pixels, ensuring that resized images maintain a balanced visual quality suitable for various digital platforms. eq. (\ref{eq:pixelcv}) shows the generation of new image for each respective pixel:
\begin{equation}\label{eq:pixelcv}
\begin{aligned}
\text{image\_pixel(x, y)} = 
& (1 - \alpha)(1 - \beta) \cdot \text{source\_image}(x1, y1) \\
& + \alpha(1 - \beta) \cdot \text{source\_image}(x2, y1) \\
& + (1 - \alpha)\beta \cdot \text{source\_image}(x1, y2) \\
& + \alpha\beta \cdot \text{source\_image}(x2, y2),
\end{aligned}
\end{equation}
where \(\alpha\) and \(\beta\) are interpolation weights based on the distances from the target pixel to its neighboring pixels in the original image. This interpolation method does not remove or discard pixels; instead, it calculates new pixel values to achieve a resized image while maintaining image integrity.

\subsubsection{Step Resizing}
We reduce the image size by steps. In this context, 'length' refers to both height and width, which are adjusted in proportion (using the same ratio) to preserve the overall scale of the image. To achieve this, we first calculate the target image length as in the following equation: 
\begin{equation} \label{eq:target_length}
\text{target\_length} = \text{source\_length} \times \text{ratio}.
\end{equation}
Later, we determine the scaling ratio needed for each step to ensure non-linear reduction as in the following equation: 
\begin{equation} \label{eq:step_ratio}
\text{step\_ratio} = \left(\frac{\text{target\_length}}{\text{source\_length}}\right)^{\frac{1}{\text{step}}}.
\end{equation}

Since the resolution must be in whole numbers, we apply the floor function. Additionally, we apply conditional reduction as in eq. (\ref{eq:next_length}), in case the length reduces below the target length:
\begin{equation} \label{eq:next_length}
\text{next\_length} = 
\begin{cases} 
\text{target\_length}, & \text{if } i = \text{step} - 1 \\
\left\lfloor \text{source\_length} \times \text{step\_ratio} \right\rfloor, & \text{otherwise}
\end{cases}
\end{equation}
\subsubsection{Gaussian Filter}
In the context of low image captioning, experimenting with Gaussian blur effects becomes particularly significant, especially when dealing with naturally blurry images. Given its practical importance, Gaussian blurring, represented by eq. (\ref{eq:gaussian_blur}), is adopted using the OpenCV library to synthesize the dataset \cite{priyanka2020image,opencv_library}. The overall algorithm is showcased in Algorithm \ref{alg:augmentation}, where \( (x, y) \) are the coordinates relative to the center of the kernel, \( \sigma\) is the standard deviations along the x and y axes, respectively.

\begin{equation} \label{eq:gaussian_blur}
G(x, y) = \frac{1}{2\pi\sigma^2} \exp({-\frac{x^2 + y^2}{2\sigma^2}}).
\end{equation}

\begin{algorithm}
\caption{Step Resizing with Gaussian Blur}\label{alg:augmentation}
\begin{algorithmic}[1]
\REQUIRE \( \text{source\_image},\text{source\_width},\text{source\_height}, \text{step}, \text{ratio} \)
\STATE \( \text{image} \gets \text{cv2.GaussianBlur(source\_image)} \)
\STATE \( \text{target\_width} \gets \text{source\_width} \times \text{ratio} \)
\STATE \( \text{target\_height} \gets \text{target\_height} \times \text{ratio} \)
\STATE \( \text{width\_ratio} \gets (\text{target\_width} / \text{source\_width})^{(1 / \text{step})} \)
\STATE \( \text{height\_ratio} \gets (\text{target\_height} / \text{source\_height})^{(1 / \text{step})} \)

\FOR{\( i = 0 \) \TO \( \text{step} - 1 \)}
    \STATE \( \text{next\_width} \gets \left\lfloor \text{image.width} \times 
    \text{width\_ratio} \right\rfloor \)
    \STATE \( \text{next\_height} \gets \left\lfloor \text{image.height} \times \text{height\_ratio} \right\rfloor \)

    \IF{\( i = \text{step} - 1 \)}
        \STATE \( \text{next\_width} \gets \text{target\_width} \)
        \STATE \( \text{next\_height} \gets \text{target\_height} \)
    \ENDIF
    
    \STATE \( \text{image} \gets \text{cv2.resize}(\text{image}, \text{next\_width}, \text{next\_height}) \)
    
    \COMMENT{Stop resizing if width reaches target resolution} \IF{\( \text{next\_width} = \text{target\_width} \) \AND \( \text{next\_height} = \text{target\_height} \)}
        \STATE \textbf{break}
    \ENDIF
\ENDFOR
\RETURN \( \text{image} \)
\end{algorithmic}
\end{algorithm}

\subsection{Model Architecture}
\begin{figure}[ht]
    \centering
\includegraphics[width=\linewidth]{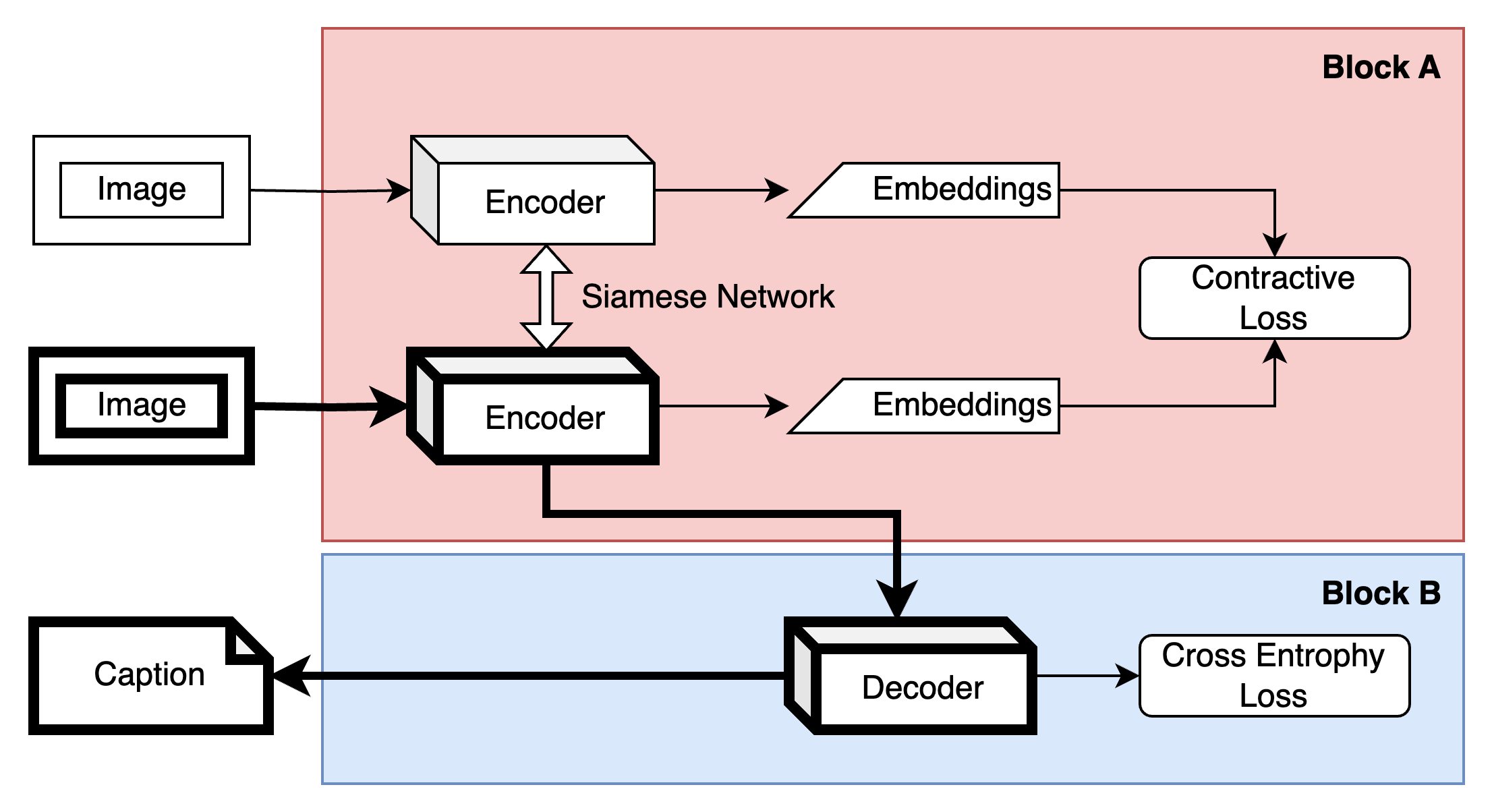}
    \caption{A depiction of the proposed architecture for SOLI: Siamese-Driven Optimization for Low-Resolution Image Latent Embedding for Captioning.}
    \label{fig:architecture}
\end{figure}

The fundamental image captioning pipeline consists of an encoder and a decoder (see Fig. \ref{fig:architecture}). The encoder generates a latent embedding, while the decoder generates a caption. The classical training pipeline uses cross-entropy loss to train as shown in Block B of Fig. \ref{fig:architecture}. The embedding is fed to the decoder and compared with labeled caption in the dataset, then it is fed for cross-entropy loss denoted as $L_{cross\_entropy}$: 
\begin{equation}\label{eq:celoss}
L_{\text{cross\_entropy}} = -\sum_{i=1}^{N} \sum_{j=1}^{M} y_{ij} \log(\hat{y}_{ij}),
\end{equation}
where \(N\) is the number of samples, \(M\) is the number of classes, \(y_{ij}\) is a binary indicator (0 or 1) of label \(j\) is the correct classification for sample \(i\), and \(\hat{y}_{ij}\) is the predicted probability that sample \(i\) belongs to class \(j\) \cite{maru2021comparison}.

Beyond that, we develop a Siamese Network with contrastive loss as shown in eq. (\ref{eq:similarity}), where \( y_i \) indicates whether a pair is considered similar (i.e., 1) or dissimilar (i.e., 0), \( m \) is the margin hyperparameter specifying the threshold distance, and \( D_i \) represents the Euclidean distance between embeddings encoded by the encoder as defined in eq. (\ref{eq:similarity}) \cite{chicco2021siamese}.

\begin{equation*}\label{eq:ctloss}
L_{\text{contrastive}} = \frac{1}{N} \sum_{i=1}^{N} \left[ y_i D_i^2 + (1 - y_i) \max(0, m - D_i)^2 \right],
\end{equation*}
\begin{equation}\label{eq:similarity}
 D_i = \| \text{encode}(d_i^{(a)}) - \text{encode}(d_i^{(b)}) \|.
\end{equation}

The overall pipeline (in \textbf{bold}) is shown in Fig. \ref{fig:architecture}. Our goal is to ensure that the produced embeddings remain consistent across all generated augment image embeddings. To achieve this, we propose SOLI (Siamese-driven Optimization for Low-Resolution Images), a multitask semi-self-supervised learning approach.
We first train the encoder using the contrastive loss (Block A) and subsequently fine-tune it with a conventional approach - cross entropy loss (Block B).

Additionally, we propose combining the losses. As illustrated in Fig. \ref{fig:architecture}, for each input from dataset \( d_n \), we obtain two images \( d_{n}^{(a)} \) and \( d_{n}^{(b)} \) with their respective labels \( y_{n}^{(a)} \) and \( y_{n}^{(b)} \), along with a similarity label \( y_{n}^{(s)} \). Both images are fed into the encoder in a Siamese Network configuration. During the training process of the Siamese network (in Block A), both the original and augmented datasets are utilized. Each image is selected individually, and a random function with a probability of 0.5 determines whether the pair is positive or negative. For positive pairs, the augmented image corresponding to the original image is selected, while for negative pairs, a random image distinct from the original is chosen. 
As shown in eq. (\ref{eq:soli}), where \( \gamma \) and \( \lambda \) are parameters added to control the weightage of the influences.

\begin{equation}\label{eq:soli}
L_{\text{SOLI}} = \gamma \cdot L_{\text{contrastive}} + \lambda \cdot L_{\text{cross\_entropy}}.
\end{equation}

We designed the following experiments to evaluate SOLI:
\begin{enumerate}
    \item \textbf{Normal Flickr8k Dataset + Classical Image Captioning Pipeline}: To evaluate the effect of LRI on the model trained in the normal pipeline.
    \item \textbf{Normal \& Augmented Dataset + Classical Image Captioning Pipeline}: To establish a baseline marker for evaluation by fine-tuning the model again on the augmented dataset.
    \item \textbf{Normal \& Augmented Dataset + Proposed SOLI appraoches}: To validate the proposed model architecture.
\end{enumerate}

Along with that, we studied 3 SOLI fine-tuning approaches: 
\begin{enumerate}
    \item \textbf{Encoder fine-tuning (SOLI-half)}, where only encoder (Block A) is trained using eq. (\ref{eq:celoss});
    
    \item \textbf{Parallel fine-tuning (SOLI-par)}, involving both encoder (Block A) and decoder (Block B) using eq. (\ref{eq:soli}); 
    
    \item \textbf{Concurrent fine-tuning (SOLI-con)}, where initially encoder (Block A) is fine-tuned using eq. (\ref{eq:celoss}), followed by fine-tuning decoder (Block B) using eq. (\ref{eq:soli}). 
\end{enumerate}
\subsection{Evaluation Metrics}
Since there are numerous evaluation metrics available, we have chosen to include two popular metrics to assess the performance of our model \cite{math11041006}.
\subsubsection{BLEU (Bilingual Evaluation Understudy)}
BLEU (B) is a widely-used metric for assessing the quality of machine-generated text by comparing its n-grams with reference translations. In this paper, we apply the popular BLEU-1 (B-1) and BLEU-4 (B-4) metrics. BLEU's simplicity and efficiency make it popular and correlate well with evaluation quality. However, BLEU (B) scores can be sensitive to text length and lack sensitivity to syntactic correctness \cite{ghandi2023deep,math11041006}.

\subsubsection{METEOR (Metric for Evaluation of Translation with Explicit ORdering)}
METEOR (M) often evaluates machine-translated texts by considering not only exact word matches but also stem matches and synonymy. It provides a robust evaluation metric that aligns better with human judgments at the segment level. Recently, it has become popular in image captioning due to criticism of BLEU (B) \cite{xu2015show, ghandi2023deep,math11041006}.

\section{Results and Discussion}
We investigated the performance of image captioning using classical encoder-decoder models and transformer-based models on the augmented datasets. 

\subsection{Performance Study on the Effect of LRI}
To ensure the validity of the experiment, we first trained the model on the normal dataset. Table \ref{tab:base} tabulates the performance when evaluating on different datasets using different combination of encoder and decoder \cite{subangkar,kumar2022imagecaptioning,cnn10543514,wolf-etal-2020-transformers}. We experimented with combinations of encoders, including VGG, ResNet, and Visual Transformer, and decoders such as LSTM-GloVe with/without Attention, and GPT Transformer. Due to variations in hyperparameter configurations, slight differences in results were observed. To ensure a fair comparison, we kept the model hyperparameters fixed throughout this research. 

As shown, the performance is reduced, significant on Dataset R0.2S50, Dataset R0.1S50, and tremendously reduced on Dataset R0.05S50, with reductions of up to 0.10. It is worth mentioning that since the encoder receives $224\times224$ as input, a scale of 0.5 from the original image does not affect the results as the dataset is $500\times500$, which is almost similar to the suggested resize. Hence, Dataset R0.5S1 obtains similar or even higher results than the normal dataset.

\begin{figure}[ht]
    \centering
    \includegraphics[width=\linewidth]{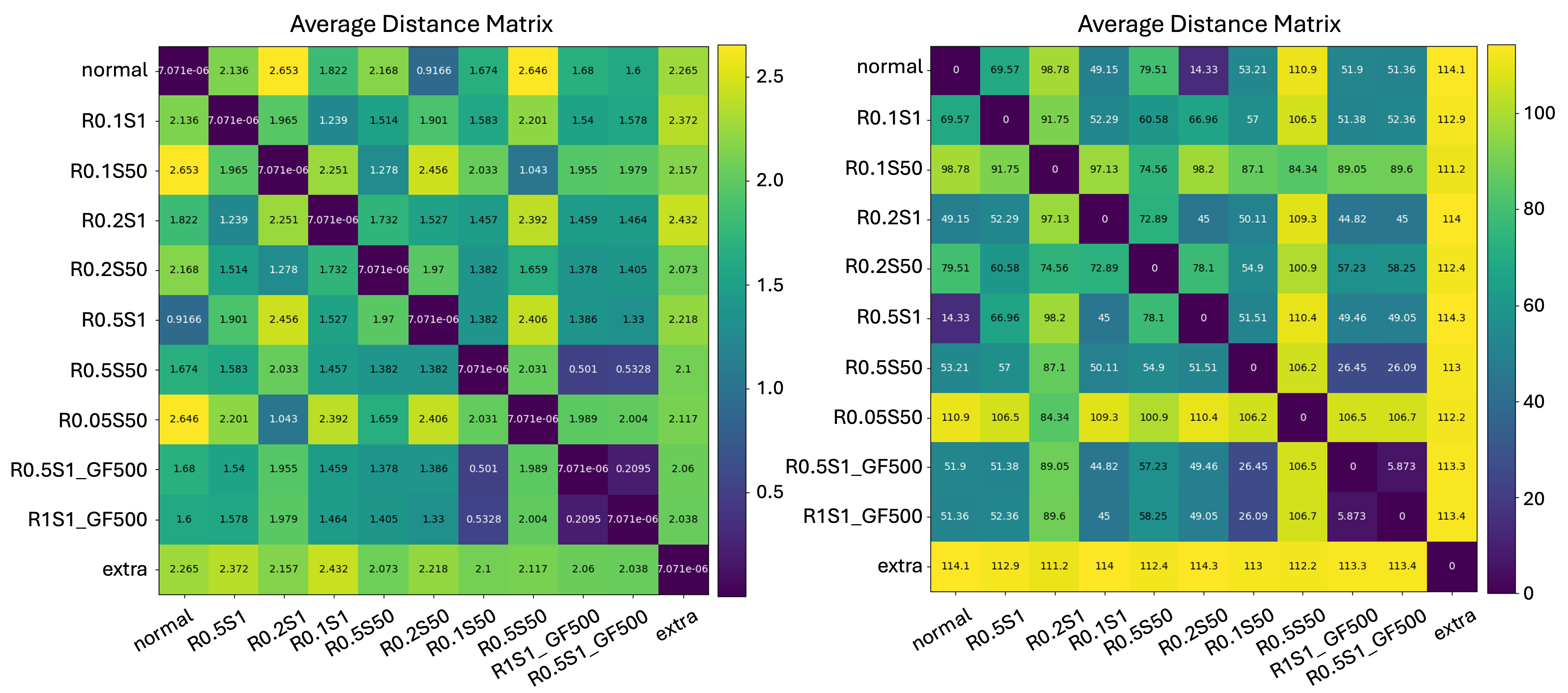}
    \caption{Similarity differences of latent embeddings from the encoder model. Left: ResNet101 Model, Right: VIT Model}
    \label{fig:encoder-normal}
\end{figure}

\begin{table*}[htbp]
    \centering
    \caption{Performance comparison in image captioning using combinations of VGG, ResNet, LSTM-GloVe, Attention, Visual Transformer, and GPT Transformer.}
    \label{tab:base}
    \begin{tabular}{|c|ccc|ccc|ccc|ccc|}
    
        \hline
        \multirow{2}{*}{\textbf{Datasets}} & \multicolumn{3}{c|}{\textbf{VGG + LSTM\_GloVe}} & \multicolumn{3}{c|}{\textbf{ResNet + LSTM\_GloVe}} & \multicolumn{3}{c|}{\textbf{ResNet+Att-LSTM-GloVe}} & \multicolumn{3}{c|}{\textbf{VIT + GPT}} \\
        \cline{2-13}
         & B1 & B4 & M & B1 & B4 & M & B1 & B4 & M & B1 & B4 & M \\
        \hline
    normal & 0.5355 & 0.1658 & \textbf{0.1938} & 0.5639 & 0.1914 & 0.2202 & 0.6053 & 0.2315 & 0.2617 & \textbf{0.7742} & 0.6909 & 0.5745 \\
     R0.5S50 & 0.5327 & 0.1592 & 0.1811 & 0.5596 & 0.1880 & 0.2065 & 0.6015 & 0.2305 & 0.2485 & 0.7638 & 0.6892 & 0.5705 \\
     R0.5S1 & \textbf{0.5393} & \textbf{0.1700} & 0.1932 & \textbf{0.5662} & \textbf{0.1957} & \textbf{0.2300} & \textbf{0.6076} & \textbf{0.2404} & \textbf{0.2622} & 0.7723 & \textbf{0.6937} & \textbf{0.5762} \\
     R0.2S50 & 0.5126 & 0.1445 & 0.1569 & 0.5386 & 0.1734 & 0.1861 & 0.5804 & 0.2183 & 0.2288 & 0.7479 & 0.6628 & 0.5665 \\
     R0.2S1 & 0.5299 & 0.1548 & 0.1609 & 0.5550 & 0.1820 & 0.1901 & 0.5993 & 0.2265 & 0.2310 & 0.7542 & 0.6912 & 0.5721 \\
    R0.1S50 & 0.5165 & 0.1460 & 0.1556 & 0.5456 & 0.1729 & 0.1839 & 0.5882 & 0.2174 & 0.2283 & 0.7284 & 0.6454 & 0.5662 \\
    R0.1S1 & 0.5266 & 0.1570 & 0.1402 & 0.5518 & 0.1850 & 0.1671 & 0.5928 & 0.2277 & 0.2096 & 0.7629 & 0.6799 & 0.5728 \\
     R1S1\_GF500 & 0.5331 & 0.1575 & 0.1400 & 0.5587 & 0.1840 & 0.1696 & 0.6029 & 0.2274 & 0.2142 & 0.7665 & 0.6819 & 0.5730 \\
     R0.5S1\_GF500 & 0.5344 & 0.1563 & 0.1405 & 0.5602 & 0.1821 & 0.1703 & 0.6055 & 0.2238 & 0.2140 & 0.7684 & 0.6844 & 0.5735 \\
    R0.05S50 & 0.4281 & 0.0556 & 0.0386 & 0.4906 & 0.1231 & 0.1165 & 0.5324 & 0.1666 & 0.1569 & 0.6814 & 0.6050 & 0.5454 \\
        \hline
    \end{tabular}
\end{table*}
Upon further analysis of the similarity differences from the latent embedding output of the encoder model, as illustrated in Fig. \ref{fig:encoder-normal}. We selected five random images and investigated the relationships among their augmented versions. Additionally, we included another random image (extra) to serve as a control group for distance reference. The average results are plotted for analysis.

From Table \ref{tab:base}, we observe that poorer results typically coincide with larger deviations from the normal dataset (e.g. R0.1S50) especially in visual transformer, leading to information loss that impacts the decoder's ability to generate accurate captions. Moreover, we also observe minimal variation in the Meteor score across different augmented datasets. This is because GPT-2 is primarily a text-based language model with exceptional grammatical fluency. As a result, the model is unaffected by the encoder input, enabling it to produce reasonable sentence structures.

\subsection{Baseline Performance}
We fine-tuned the attention model on Dataset R0.2S50 and Dataset R0.1S50 without resetting weights. Table \ref{tab:datafineeach} shows improved performance, albeit with a slight decrease on other datasets (other than them, e.g. Normal dataset). These results suggest effective refinement using augmented datasets.

\begin{table*}[htbp]
    \centering
    \caption{Performance comparison in image captioning after fine-tuning in Dataset R0.2S50 and Dataset R0.1S50.}
    \label{tab:datafineeach}
    \begin{tabular}{|c|ccc|ccc|ccc|ccc|}
    
        \hline
    \multicolumn{1}{|c|}{\multirow{3}[0]{*}{\textbf{Datasets}}} & \multicolumn{6} {c|}{\textbf{ResNet+Att-LSTM-GloVe}}& \multicolumn{6}{|c|}{\textbf{VIT + GPT}}\\
    \cline{2-13}
      & \multicolumn{3}{c|}{\textbf{Fine Tune on R0.2S50}} & \multicolumn{3}{c|}{\textbf{Fine Tune on R0.1S50}} & \multicolumn{3}{c|}{\textbf{Fine Tune on R0.2S50}} & \multicolumn{3}{c|}{\textbf{Fine Tune on R0.1S50}} \\
    \cline{2-13}
     & B1 & B4 & M & B1 & B4 & M & B1 & B4 & M & B1 & B4 & M \\
    \hline
           normal & 0.5822 & 0.2078 & 0.2294 & 0.5845 & 0.2026 & 0.2357 & 0.7222 & 0.6491 & 0.5744 & 0.7207 & 0.6479 & 0.5417 \\
     R0.5S50 & 0.5774 & 0.2035 & 0.2251 & 0.5798 & 0.2046 & 0.2264 & 0.7239 & 0.6616 & 0.5709 & 0.7255 & 0.6591 & 0.5688 \\
     R0.5S1 & 0.5819 & 0.2116 & 0.2298 & 0.5791 & 0.2174 & 0.2385 & 0.7234 & 0.6596 & 0.5704 & 0.7200 & 0.6606 & 0.5698 \\
     \textbf{R0.2S50} & \textbf{0.6028} & \textbf{0.2341} & \textbf{0.2496} & 0.5506 & 0.1894 & 0.2043 & \textbf{0.7431} & \textbf{0.6673} & \textbf{0.5720} & 0.7168 & 0.6600 & 0.5621 \\
     R0.2S1 & 0.5726 & 0.1982 & 0.2056 & 0.5780 & 0.2025 & 0.2011 & 0.7232 & 0.6597 & 0.5693 & 0.7219 & 0.6598 & 0.5675 \\
    \textbf{R0.1S50} & 0.5613 & 0.1912 & 0.1993 & \textbf{0.6021} & \textbf{0.2352} & \textbf{0.2495} & 0.7015 & 0.6400 & 0.5713 & \textbf{0.7407} & \textbf{0.6675} & \textbf{0.5734} \\
    R0.1S1 & 0.5671 & 0.2042 & 0.1867 & 0.5710 & 0.2050 & 0.1866 & 0.7196 & 0.6590 & 0.5676 & 0.7219 & 0.6585 & 0.5670 \\
     R1S1\_GF500 & 0.5758 & 0.2017 & 0.1894 & 0.5802 & 0.2066 & 0.1948 & 0.7209 & 0.6569 & 0.5678 & 0.7189 & 0.6577 & 0.5671 \\
     R0.5S1\_GF500 & 0.5771 & 0.2010 & 0.1881 & 0.5855 & 0.2052 & 0.1913 & 0.7212 & 0.6589 & 0.5668 & 0.7222 & 0.6556 & 0.5606 \\
    R0.05S50 & 0.5187 & 0.1491 & 0.1321 & 0.5105 & 0.1455 & 0.1325 & 0.6947 & 0.6036 & 0.5502 & 0.6976 & 0.6110 & 0.5488 \\
        \hline
    \end{tabular}
\end{table*}

Later, we fine-tuned the model using the entire augmented dataset. Since this is the common way to handle LRI problem, the results (averaged) presented in Table \ref{tab:datafine} will serve as a baseline for future experiments. It's important to note that we excluded image scaling with a factor of 0.5. This is because such scaling significantly reduced the information content in the images, leading to performance exceeding human ability which is an unrealistic benchmark. Generally, the performance is slightly better than when only fine-tuning on a single dataset, but overall, it still decreased.

\begin{table}[htbp]
    \centering
    \caption{Performance comparison in image captioning after fine-tuning using all normal and augmented dataset}
    \label{tab:datafine}\
    \setlength{\tabcolsep}{4pt}
    \begin{tabular}{|c|ccc|ccc|}
        \hline
        \multirow{2}{*}{\textbf{Datasets}} & \multicolumn{3}{c|}{\textbf{ResNet+Att-LSTM-GloVe}} & \multicolumn{3}{c|}{\textbf{VIT +GPT}} \\
        \cline{2-7}
        \
        & B1 & B4 & M & B1 & B4 & M \\
        \hline
        normal & 0.5770 & 0.1987 & 0.2324 & 0.7133 & 0.6210 & 0.5615 \\
     R0.5S50 & 0.5715 & 0.2020 & 0.2224 & 0.7100 & 0.6213 & 0.5578 \\
     R0.5S1 & 0.5773 & 0.2105 & 0.2385 & 0.7103 & 0.6311 & 0.5596 \\
     R0.2S50 & 0.5656 & 0.1851 & 0.2299 & 0.7100 & 0.6254 & 0.5580 \\
     R0.2S1 & 0.5725 & 0.1959 & 0.2200 & 0.7121 & 0.6215 & 0.5570 \\
    R0.1S50 & 0.5705 & 0.2010 & 0.2226 & 0.7105 & 0.6264 & 0.5575 \\
    R0.1S1 & 0.5668 & 0.1994 & 0.2109 & 0.7218 & 0.6276 & 0.5595 \\
     R1S1\_GF500 & 0.5697 & 0.2091 & 0.2186 & 0.7149 & 0.6200 & 0.5574 \\
     R0.5S1\_GF500 & 0.5826 & 0.2031 & 0.2172 & 0.7181 & 0.6224 & 0.5577 \\
    R0.05S50 & 0.5153 & 0.1593 & 0.1615 & 0.6864 & 0.5833 & 0.5347 \\
    \hline
    \textbf{Mean} & \textbf{0.5726} & \textbf{0.2005} & \textbf{0.2236} & \textbf{0.7134} & \textbf{0.6241} & \textbf{0.5584} \\
        \hline
    \end{tabular}
\end{table}

\subsection{Performance of the Proposed SOLI Method }

As in Table \ref{tab:soli}, the overall performance increased, especially for SOLI-par. The SOLI-half performance justified fine-tuning only the encoder can bring better captionign result. However, both methods, especially SOLI-half, significantly degrade the model's performance on original images. 

The BLEU-4 (B4) score for the ResNet+Att-LSTM-GloVe model increased from 0.2055 to 0.2181 (improvement: 0.0126), and for the VIT+GPT model, it increased from 0.6241 to 0.6536 (improvement: 0.0387).
This is expected not to outperform the ceiling result (the fine-tuned result in Table \ref{tab:datafineeach}) as it only aims to approach it. Using the same example, they still have small gaps towards their optimum ($<0.2352$, $0.0171$ to ceiling) and ($<0.6628$, $0.0295$ to ceiling) respectively. Nevertheless, SOLI-con did not yield significantly different results compared to SOLI-par and is not presented here.

\begin{figure}[ht]
    \centering
    \includegraphics[width=\linewidth]{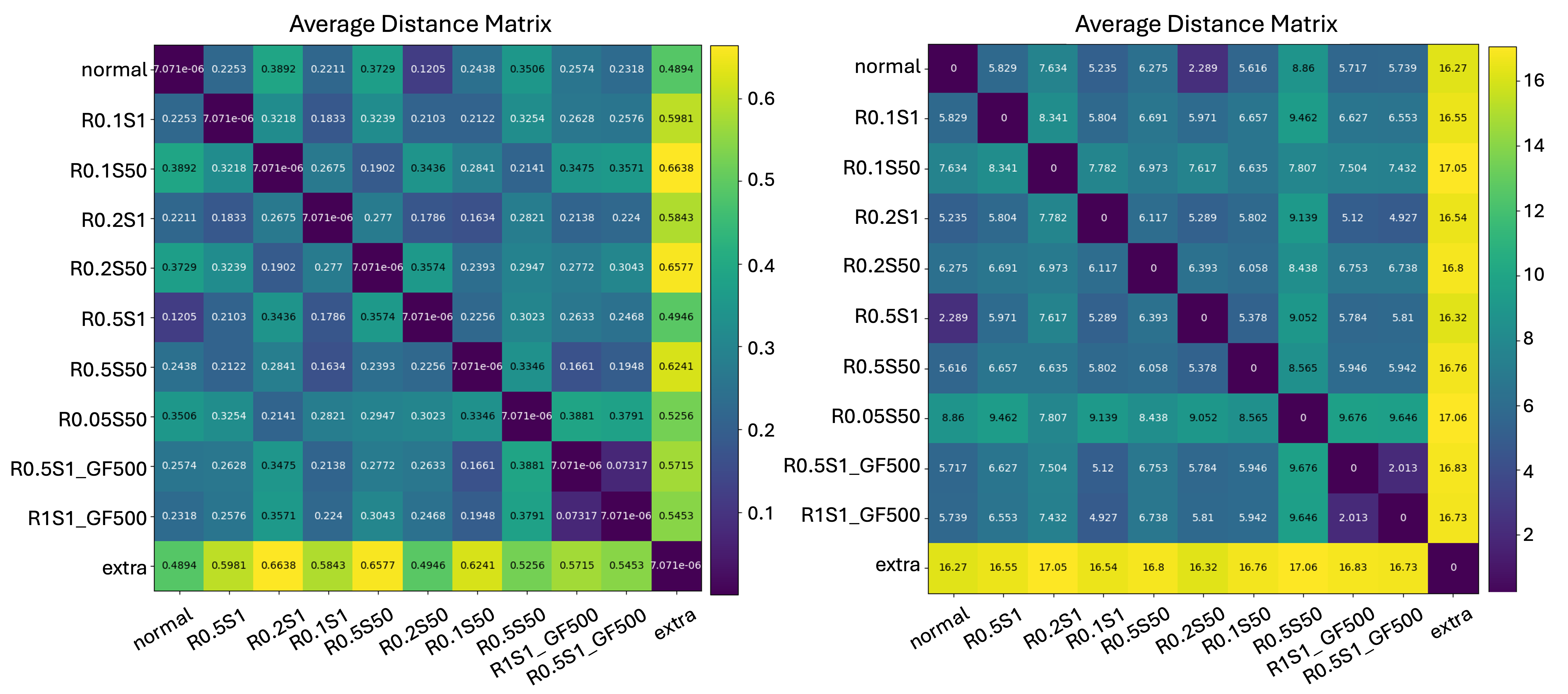}
    \caption{Similarity differences in latent embeddings from the encoder model after fine-tuning (SOLI-par). Left: ResNet101 Model, Right: VIT Model}
    \label{fig:ensiamese}
\end{figure}

This SOLI-par result is supported by Fig. \ref{fig:ensiamese}. As observed, the distances between each augmented latent embedding dataset are closer, while the control group indicates that the model continues to effectively differentiate between images.

\begin{table*}[htbp]
    \centering
    \caption{Performance comparison in image captioning with proposed SOLI approaches using all normal and augmented dataset}
    \label{tab:soli}
    \begin{tabular}{|c|ccc|ccc|ccc|ccc|}
    
        \hline
    \multicolumn{1}{|c|}{\multirow{3}[0]{*}{\textbf{Datasets}}} & \multicolumn{6} {c|}{\textbf{ResNet+Att-LSTM-GloVe}}& \multicolumn{6}{|c|}{\textbf{VIT + GPT}}\\
    \cline{2-13}
      & \multicolumn{3}{c|}{\textbf{SOLI-half}} & \multicolumn{3}{c|}{\textbf{SOLI-par}} & \multicolumn{3}{c|}{\textbf{SOLI-half}} & \multicolumn{3}{c|}{\textbf{SOLI-par}} \\
    \cline{2-13}
     & B1 & B4 & M & B1 & B4 & M & B1 & B4 & M & B1 & B4 & M \\
    \hline
        normal & 0.5825 & 0.2038 & 0.2228 & 0.5966 & 0.2187 & 0.2436 & 0.7206 & 0.6312 & 0.5610 & 0.7367 & 0.6470 & 0.5693 \\
     R0.5S50 & 0.5767 & 0.2070 & 0.2196 & 0.5847 & 0.2154 & 0.2339 & 0.7213 & 0.6299 & 0.5635 & 0.7350 & 0.6552 & 0.5598 \\
     R0.5S1 & 0.5823 & 0.2159 & 0.2287 & 0.5992 & 0.2243 & 0.2498 & 0.7269 & 0.6307 & 0.5625 & 0.7393 & 0.6567 & 0.5654 \\
     R0.2S50 & 0.5706 & 0.1903 & 0.2201 & 0.5790 & 0.2086 & 0.2311 & 0.7174 & 0.6284 & 0.5622 & 0.7304 & 0.6526 & 0.5625 \\
     R0.2S1 & 0.5777 & 0.2011 & 0.2152 & 0.5857 & 0.2196 & 0.2317 & 0.7197 & 0.6286 & 0.5625 & 0.7348 & 0.6462 & 0.5641 \\
    R0.1S50 & 0.5760 & 0.2063 & 0.2226 & 0.5840 & 0.2128 & 0.2343 & 0.7186 & 0.6312 & 0.5660 & 0.7312 & 0.6541 & 0.5652 \\
    R0.1S1 & 0.5721 & 0.2046 & 0.2212 & 0.5846 & 0.2184 & 0.2317 & 0.7185 & 0.6275 & 0.5616 & 0.7333 & 0.6555 & 0.5619 \\
     R1S1\_GF500 & 0.5747 & 0.2146 & 0.2231 & 0.5833 & 0.2222 & 0.2305 & 0.7210 & 0.6273 & 0.5616 & 0.7332 & 0.6562 & 0.5620 \\
     R0.5S1\_GF500 & 0.5880 & 0.2084 & 0.2224 & 0.5962 & 0.2225 & 0.2320 & 0.7185 & 0.6276 & 0.5616 & 0.7323 & 0.6515 & 0.5612 \\
    R0.05S50 & 0.5205 & 0.1644 & 0.1616 & 0.5291 & 0.1732 & 0.1729 & 0.6943 & 0.6012 & 0.5425 & 0.7003 & 0.6335 & 0.5442 \\
    \hline
    \textbf{Mean}  & \textbf{0.5779} & \textbf{0.2058} & \textbf{0.2217} & \textbf{0.5881} & \textbf{0.2181} & \textbf{0.2354} & \textbf{0.7203} & \textbf{0.6292} & \textbf{0.5625} & \textbf{0.7340} & \textbf{0.6536} & \textbf{0.5635} \\
    \hline
    \end{tabular}
\end{table*}

\section{Conclusion}
This paper investigates and demonstrates the feasibility of using the proposed SOLI approach to enhance performance on low-resolution images. The findings provide evidence supporting the effectiveness of this approach in improving image processing outcomes for low resolution images. 

Moving forward, future research will explore incremental learning methodologies, including the integration of reinforcement learning techniques. Additionally, there will be a focus on evaluating the trade-off between training/inference costs and accuracy, aiming to achieve efficient and effective deployment of the developed models in practical applications. 

The demonstration and resources, including the code repository, can be found at this link: \textcolor{blue}{\url{https://imgcap.jingjietan.com/}}.


\section*{Acknowledgment}
The authors gratefully acknowledge the support of Grid5000 for providing computational resources and the Embassy of France to Malaysia Doctoral Research Mobility Grant, which facilitated the research collaboration between Université  Sorbonne Paris Nord and Universiti Tunku Abdul Rahman. 



\bibliographystyle{IEEEtran}
\bibliography{references}




\end{document}